# Nondeterministic fuzzy automata

Yongzhi Cao[a,b], Yoshinori Ezawa[c]

[a]*Institute of Software, School of Electronics Engineering and Computer Science, Peking University, Beijing 100871, China*
[b]*Key Laboratory of High Confidence Software Technologies (Peking University), Ministry of Education, China*
[c]*Faculty of Informatics, Kansai University, Osaka 569-1095, Japan*

## Abstract

Fuzzy automata have long been accepted as a generalization of nondeterministic finite automata. A closer examination, however, shows that the fundamental property—nondeterminism—in nondeterministic finite automata has not been well embodied in the generalization. In this paper, we introduce nondeterministic fuzzy automata with or without $\epsilon$-moves and fuzzy languages recognized by them. Furthermore, we prove that (deterministic) fuzzy automata, nondeterministic fuzzy automata, and nondeterministic fuzzy automata with $\epsilon$-moves are all equivalent in the sense that they recognize the same class of fuzzy languages.

## 1. Introduction

Finite automata (also called finite-state automata or finite-state machines) are probably best known computational devices and are used to model operations of many systems in practice. Automata theory is closely related to formal language theory as the automata are often classified by the class of formal languages they are able to recognize [15]. A finite automaton gives a finite representation of a regular language that may be an infinite set. It is widely known that finite automata are significant in many different areas, including computer science, linguistics, mathematics, electrical engineering, philosophy, and biology.

To deal with imprecision due to fuzziness in modeling some systems, fuzzy automata and fuzzy languages have been proposed as a reasonable extension of classical automata and formal language theory. The mathematical formulation of fuzzy automata was first proposed by Wee in 1967 [36]. The basic idea in the formulation is that, unlike the classical case, a fuzzy automaton can switch from one state to another one to a certain possibility degree, and thus it is capable of capturing the uncertainty appearing in states or state transitions of a system. A fuzzy language over an alphabet $\Sigma$, originally introduced in [18, 25], is defined as a fuzzy subset of $\Sigma^*$, where $\Sigma^*$ stands for the Kleene closure of $\Sigma$. In this sense, a fuzzy language $\mathcal{L}$ allows for strings in $\Sigma^*$ that are not completely in or out the language; rather each string $s$ has a membership grade $\mathcal{L}(s)$, which measures its degree of membership in $\mathcal{L}$.

In the literature up to now, various variants of fuzzy automata have been proposed in different modeling situations (see, for example, [2, 6, 7, 8, 9, 10, 16, 17, 22, 23, 26, 27, 28, 33, 38]) and the notions of fuzzy automata and fuzzy languages have proved useful in many areas [1, 4, 11, 12, 13, 20, 21, 30, 31, 32, 34, 35, 37, 39]. In terms of fuzzy transition functions, fuzzy automata may be broadly classified into three types: The first type [1, 3, 6, 7, 9, 11, 12, 13, 16, 21, 22, 23, 27, 28, 32, 33, 35, 37, 39] uses fuzzy transition functions like $\delta : Q \times \Sigma \longrightarrow \mathcal{F}(Q)$, where $Q$ represents the state set, $\Sigma$ is the input alphabet, and $\mathcal{F}(Q)$ is the set of all fuzzy subsets of $Q$. Note that such a fuzzy transition function can be equivalently converted into $\delta' : Q \times \Sigma \times Q \longrightarrow [0, 1]$ and can also be represented by fuzzy states and fuzzy transition matrices. Frequently the interval $[0, 1]$ has been replaced by a lattice or other suitable algebraic structure. The second type [5, 24, 34] employs the same transition function as that of a deterministic finite automaton and encodes the fuzziness by fuzzy initial state(s) and fuzzy final states. The third type [4, 8, 10, 17, 20, 38] adds more parameters or components to the first type.

Although fuzzy automata include nondeterministic finite automata as a special case, it is worth noting that to the best of our knowledge, all fuzzy transition functions of the fuzzy automata in the existing literature are deterministic in the sense that an automaton in its current state and with the symbol currently read gives rise to a unique next (crisp or fuzzy) state. It turns out that the fundamental property—nondeterminism—in nondeterministic finite automata has not been fully embodied in the extension from finite automata to fuzzy automata. As is well known, nondeterminism is, however, essential for modeling scheduling freedom, implementation freedom, the external environment, and incomplete information (see, for example, [14]). In fact, nondeterminism implicitly appears in the earlier work of Wee and Fu [37], where they formulated a learning composite fuzzy automaton operating alternatively between next two states. Unfortunately, nondeterminism has been largely ignored in the area of fuzzy automata over the past 40 years.

The present paper is mainly concerned with fuzzy automata with nondeterminism. The main driving idea for it is that the presence of fuzziness should not hide the underlying nondeterminism, but rather gives more information. We introduce nondeterministic fuzzy automata, nondeterministic fuzzy automata with $\epsilon$-moves, and fuzzy languages recognized by them. Note that the term of nondeterministic fuzzy automata has already been used by some authors (for example, [3, 19, 22]), but it takes on a different meaning. In fact, it corresponds exactly to the notion of deterministic fuzzy automata here. Following the study of nondeterministic finite automata [29], we show that conventional fuzzy automata, nondeterministic fuzzy automata, and nondeterministic fuzzy automata with $\epsilon$-moves are all equivalent in the sense that they recognize the same class of fuzzy languages.

The remainder of this paper is structured as follows. We review some basic facts on fuzzy sets, fuzzy automata, and fuzzy languages in Section 2. After a closer examination of nondeterministic finite automata, the notions of nondeterministic fuzzy automata and fuzzy languages accepted by them are introduced in Section 3 and are extended to nondeterministic fuzzy automata with $\epsilon$-moves in Section 4. We establish the equivalence among deterministic fuzzy automata, nondeterministic fuzzy automata, and nondeterministic fuzzy automata with $\epsilon$-moves in Section 5 and conclude the paper in Section 6.

## 2. Preliminaries

In this section, we briefly recall a few basic facts on fuzzy sets, fuzzy automata, and fuzzy languages.

Let us begin with the notion of fuzzy sets initiated by Zadeh [40]. Suppose that $X$ is a universal set. A *fuzzy set $A$*, or rather a *fuzzy subset $A$* of $X$, is defined by a function assigning to each element $x$ of $X$ a value $A(x)$ in the real unit closed interval $[0, 1]$. Such a function is called a *membership function*, which is a generalization of the characteristic function associated to a crisp subset of $X$. The value $A(x)$ characterizes the degree of membership of $x$ in $A$. In light of this, a fuzzy subset of $X$ can be viewed as a possibility distribution on $X$.

We denote by $\mathcal{F}(X)$ the set of all fuzzy subsets of $X$ and by $\mathcal{P}(X)$ the power set of $X$. For any $A, B \in \mathcal{F}(X)$, we say that $A$ is contained in $B$ (or $B$ contains $A$), denoted by $A \subseteq B$, if $A(x) \leq B(x)$ for all $x \in X$. We say that $A = B$ if and only if $A \subseteq B$ and $B \subseteq A$.

The *support* of a fuzzy set $A$ is a crisp set defined by

$$\mathrm{supp}(A) = \{x \in X \mid A(x) > 0\}.$$

A fuzzy set is said to be *empty* if its support is empty, that is, its membership function is identically zero on $X$. We use $\Phi$ to denote the empty fuzzy set. Whenever $\mathrm{supp}(A)$ is finite, say $\mathrm{supp}(A) = \{x_1, x_2, \ldots, x_n\}$, we may write $A$ in Zadeh's notation as

$$A = \frac{A(x_1)}{x_1} + \frac{A(x_2)}{x_2} + \cdots + \frac{A(x_n)}{x_n}.$$

With this notation, $1/x$ is a singleton in $X$, i.e., the fuzzy subset of $X$ with membership 1 at $x$ and with zero membership for all the other elements of $X$.

For any family $\lambda_i$, $i \in I$, of elements of $[0, 1]$, we write $\vee_{i \in I} \lambda_i$ or $\vee \{\lambda_i \mid i \in I\}$ for the supremum of $\{\lambda_i \mid i \in I\}$, and $\wedge_{i \in I} \lambda_i$ or $\wedge \{\lambda_i \mid i \in I\}$ for the infimum. In particular, if $I$ is finite, then $\vee_{i \in I} \lambda_i$ and $\wedge_{i \in I} \lambda_i$ are the greatest element and the least element of $\{\lambda_i \mid i \in I\}$, respectively. For any $A \in \mathcal{F}(X)$, the *height* of $A$ is defined as

$$\mathrm{height}(A) = \vee_{x \in X} A(x).$$

Given $A, B \in \mathcal{F}(X)$, the *union* of $A$ and $B$, denoted $A \cup B$, is defined by the membership function

$$(A \cup B)(x) = A(x) \vee B(x)$$

for all $x \in X$; the *intersection* of $A$ and $B$, denoted $A \cap B$, is given by the membership function

$$(A \cap B)(x) = A(x) \wedge B(x)$$

for all $x \in X$. Let $\lambda \in [0, 1]$ and $A \in \mathcal{F}(X)$. The *scale product* $\lambda \cdot A$ of $\lambda$ and $A$ is defined by

$$(\lambda \cdot A)(x) = \lambda \wedge A(x)$$

for every $x \in X$; this is again a fuzzy subset of $X$.

We now turn to the concept of fuzzy automata. According to the compositional inference methods used in a fuzzy automaton, one may classify fuzzy automata into several types, such as max-min automata, min-max automata, max-product automata, and so on [26]. For each fuzzy finite state automaton, there are two different, but essentially equivalent, presentations. One is to encode the fuzziness of an automaton by fuzzy state vectors and fuzzy transition matrices. In this case, both state set itself and transition function are crisp. The other is to encode the fuzziness by a fuzzy transition function. For our purpose, we pursue the latter and restrict ourselves to the max-min automaton model for simplicity.

**Definition 1.** *A fuzzy automaton $M$ is a five-tuple $(Q, \Sigma, \delta, q_0, F)$, where*

(1) *$Q$ is a finite, nonempty set of states.*
(2) *$\Sigma$ is a finite input alphabet.*
(3) *$\delta : Q \times \Sigma \longrightarrow \mathcal{F}(Q)$ is a fuzzy transition function.*
(4) *$q_0$ in $Q$ is the initial state.*
(5) *$F$ in $\mathcal{F}(Q)$ is the fuzzy set of final states.*

The fuzzy transition function $\delta$ takes a state in $Q$ and an input symbol in $\Sigma$ as arguments and returns a possibility distribution on $Q$ (i.e., fuzzy subset of $Q$). For each $q \in Q$, $F(q)$ indicates intuitively the degree to which $q$ is a final state.

The notion of this fuzzy automaton is usually based upon the following premises: The fuzzy automaton can stay in some crisp states simultaneously, to a certain possibility degree in each. Those degrees are defined by a state membership function. More concretely, for any $p, q \in Q$ and $a \in \Sigma$, we can interpret $\delta(q, a)(p)$ as the possibility degree to which the automaton in state $q$ and with input $a$ may enter state $p$. If we had encoded the fuzziness of the automaton by fuzzy state vectors and fuzzy transition matrices, the result of inputting $a$ at state $q$ would be a fuzzy state vector having those possibility degrees as components. Therefore, from this point of view, the fuzzy automaton is deterministic in the sense that the automaton in state $q$ and with input $a$ gives rise to a unique possibility distribution on $Q$, and thus, we will refer to the fuzzy automaton in Definition 1 as *deterministic fuzzy automaton*.

For clarity, we sometimes use the more suggestive notation $q \xrightarrow{a} \mu$ to denote $\delta(q, a) = \mu$. Note that the above fuzzy transition function $\delta$ is defined as a function from $Q \times \Sigma \times Q$ to $[0, 1]$ in some references. In fact, both these representations mean the same thing because in the obvious way, each function $\delta : Q \times \Sigma \longrightarrow \mathcal{F}(Q)$ is exactly corresponding to a function $\delta' : Q \times \Sigma \times Q \longrightarrow [0, 1]$.

We end this section with the notion of fuzzy languages. Denote by $\Sigma^*$ the set of all finite strings constructed by concatenation of elements of $\Sigma$, including the empty string $\epsilon$. To describe what happens when we start in any state and follow any sequence of inputs, we extend the fuzzy transition function to strings by using max-min compositional inference method.

**Definition 2.** *Let $M = (Q, \Sigma, \delta, q_0, F)$ be a deterministic fuzzy automaton.*

(1) *The extended fuzzy transition function from $Q \times \Sigma^*$ to $\mathcal{F}(Q)$, denoted by the same notation $\delta$, is defined inductively as follows:*

$$
\begin{aligned}
\delta(q, \epsilon) &= \frac{1}{q} \\
\delta(q, sa) &= \cup_{p \in Q} \left[ \delta(q, s)(p) \cdot \delta(p, a) \right]
\end{aligned}
$$

*for all $q \in Q$, $s \in \Sigma^*$, and $a \in \Sigma$, where $1/q$ is a singleton in $Q$ and $\delta(q, s)(p) \cdot \delta(p, a)$ stands for the scale product of the membership $\delta(q, s)(p)$ and the fuzzy set $\delta(p, a)$.*

(2) *The* language $\mathcal{L}(M)$ *accepted by $M$ is a fuzzy subset of $\Sigma^*$ with the membership function defined by*

$$\mathcal{L}(M)(s) = \text{height}(\delta(q_0, s) \cap F)$$

*for all $s \in \Sigma^*$. The membership $\mathcal{L}(M)(s)$ is the degree to which $s$ is accepted by $M$.*

## 3. Nondeterministic fuzzy automata

This section is devoted to the notions of nondeterministic fuzzy automata (without $\epsilon$-moves) and fuzzy languages accepted by them.

We start with the following observation. In the literature, the concept of (deterministic) fuzzy automata has long been accepted as a generalization of nondeterministic finite automata. Indeed, it is like this. What we want to point out is that the fundamental property—nondeterminism—in nondeterministic finite automata has not been fully embodied in the generalization. To see this, let us review the definition of nondeterministic finite automata.

A *nondeterministic finite automaton $M$* is a five-tuple $(Q, \Sigma, \delta, q_0, F)$, where $Q$ is a finite nonempty set of states, $\Sigma$ is a finite input alphabet, $\delta$ is a mapping of $Q \times \Sigma$ into subsets of $Q$, $q_0$ in $Q$ is the initial state, and $F \subseteq Q$ is the set of final states.

As is well known, the important difference between the nondeterministic and deterministic case is that $\delta(q, a)$ is a (possibly empty) set of states rather than a single state. For $\delta(q, a) = \{p_1, p_2, \ldots, p_k\}$, there are at least two kinds of interpretations: One is that $M$, in state $q$, scanning $a$ on its input tape, chooses any one of $p_1, p_2, \ldots, p_k$ as the next state. The other is that $M$, in state $q$, splits itself into $k$ many copies where each of them moves along one of the $k$ transitions from $q$ labeled with $a$.

One can view a nondeterministic finite automaton $M = (Q, \Sigma, \delta, q_0, F)$ as a deterministic fuzzy automaton $\widehat{M}$ in the following natural way, which might be the only one. The deterministic fuzzy automaton $\widehat{M}$ is defined as $(Q, \Sigma, \widehat{\delta}, q_0, \widehat{F})$, where the components different from those in $M$ are given by

$$\widehat{\delta}(q, a)(p) = \begin{cases} 1, & \text{if } p \in \delta(q, a) \\ 0, & \text{otherwise} \end{cases} \quad \text{and} \quad \widehat{F}(q) = \begin{cases} 1, & \text{if } q \in F \\ 0, & \text{otherwise} \end{cases}$$

for any $p, q \in Q$ and $a \in \Sigma$. In light of this, deterministic fuzzy automata are certainly a generalization of nondeterministic finite automata. It should be noted that $\widehat{M}$, in state $q$, has only one transition labeled with $a$, i.e., $q \xrightarrow{a} \widehat{\delta}(q, a)$, and thus, no nondeterminism arises.

The above observation motivates us to consider fuzzy automata with nondeterminism, formalized in the following central definition.

**Definition 3.** *A nondeterministic fuzzy automaton $M$ is a five-tuple $(Q, \Sigma, \delta, q_0, F)$, where*

(1) *$Q$ is a finite nonempty set of states.*
(2) *$\Sigma$ is a finite input alphabet.*
(3) *$\delta : Q \times \Sigma \longrightarrow \mathcal{P}(\mathcal{F}(Q))$ is a fuzzy transition function which for each $q \in Q$ and $a \in \Sigma$ gives a set $\delta(q, a)$ of possibility distributions on $Q$.*
(4) *$q_0$ in $Q$ is the initial state.*
(5) *$F$ in $\mathcal{F}(Q)$ is the fuzzy set of final states.*

We use $q \xrightarrow{a} \mu$ to denote that $\mu \in \delta(q, a)$, and refer to $q \xrightarrow{a} \mu$ as a transition. Following the interpretations of transitions in nondeterministic finite automata, for $\delta(q, a) = \{\mu_i \mid i \in I\}$, we may say that the nondeterministic fuzzy automaton $M$, in state $q$, scanning $a$ on its input tape, chooses any one of possibility distributions in $\delta(q, a)$, say $\mu_i$, and then enters every state $p$ with possibility degree $\mu_i(p)$. For each state $q$, the outgoing transitions $q \xrightarrow{a} \mu$ represent the nondeterministic alternatives in the state $q$. We may also think that $M$, in state $q$, splits itself into $|I|$[1] many copies where each of them moves along one of the $|I|$ transitions like $q \xrightarrow{a} \mu_i$.

---

[1] The notation $|I|$ denotes the cardinality of $I$, which may be infinite.

In contrast with Definition 1, the unique difference between a nondeterministic fuzzy automaton and a deterministic one lies in the fuzzy transition function. For each $a \in \Sigma$, every state $q$ in a deterministic fuzzy automaton always has exactly one transition (possibly $q \xrightarrow{a} \Phi$) labeled by $a$. In other words, a deterministic fuzzy automaton does not allow nondeterministic choices between transitions involving the same input symbol, that is, if $q \xrightarrow{a} \mu$ and $q \xrightarrow{a} \mu'$, then $\mu = \mu'$. As opposed to this, in a nondeterministic fuzzy automaton there could be states that have none, one, or more transitions labeled by the same input symbol. Clearly, deterministic fuzzy automata arise as a special case of nondeterministic fuzzy automata by identifying each fuzzy transition function $\delta : Q \times \Sigma \longrightarrow \mathcal{F}(Q)$ with a fuzzy transition function $\delta' : Q \times \Sigma \longrightarrow \mathcal{P}(\mathcal{F}(Q))$, where $\delta'$ is defined by

$$\delta'(q, a) = \{\delta(q, a)\}.$$

We now show how to view a nondeterministic finite automaton $M = (Q, \Sigma, \delta, q_0, F)$ as a nondeterministic fuzzy automaton. One way of looking at it is to view $M$ as a deterministic fuzzy automaton $\widehat{M}$ and then view $\widehat{M}$ as a nondeterministic fuzzy automaton. As argued prior to Definition 3, this way fails to transfer the nondeterminism. A better way is to view directly $M$ as a nondeterministic fuzzy automaton $M' = (Q, \Sigma, \delta', q_0, F')$, where the components different from those in $M$ are given by

$$\delta'(q, a) = \left\{ \frac{1}{p} \,\middle|\, p \in \delta(q, a) \right\} \quad \text{and} \quad F'(q) = \begin{cases} 1, & \text{if } q \in F \\ 0, & \text{otherwise} \end{cases}$$

for any $q \in Q$ and $a \in \Sigma$. For example, $\delta'(q, a) = \left\{ \frac{1}{p_1}, \frac{1}{p_2}, \ldots, \frac{1}{p_k} \right\}$ if $\delta(q, a) = \{p_1, p_2, \ldots, p_k\}$. In this way, we keep the nondeterminism in the extension from nondeterministic finite automata to nondeterministic fuzzy automata.

As usual, we can visualize a nondeterministic fuzzy automaton $M = (Q, \Sigma, \delta, q_0, F)$ by a transition diagram. The states of $M$ are circled in the diagram, and in particular, whenever $F(q) > 0$, we mark the state $q$ by a double circle and write the value $F(q)$ around the double circle. The initial state has an in-going arrow without a source. For any transition, if it has $\Phi$ as the possibility distribution, then it is trivial and thus we omit it in the diagram. Otherwise, the transition is depicted via two parts: an arrow for input symbol and a bunch of arrows for the possibility degrees of entering next states. Note that this transition diagram representation will also be applied to deterministic fuzzy automata in the sequel.

To understand the definition, let us look at an example.

**Example 1.** *Consider the nondeterministic fuzzy automaton $M = (Q, \Sigma, \delta, q_0, F)$, where $Q = \{q_0, q_1, \ldots, q_4\}$, $\Sigma = \{a, b\}$, $\delta$ is given in Table 1, and $F = \frac{0.5}{q_2} + \frac{0.9}{q_4}$. The transition diagram of $M$ is depicted in Figure 1.*

| $\delta$ | $a$ | $b$ |
|---|---|---|
| $q_0$ | $\left\{ \frac{0.9}{q_1} + \frac{0.2}{q_2}, \frac{0.2}{q_2} + \frac{0.9}{q_3} \right\}$ | $\emptyset$ |
| $q_1$ | $\emptyset$ | $\left\{ \frac{0.1}{q_1} + \frac{0.7}{q_4}, \frac{0.7}{q_2} + \frac{0.1}{q_4} \right\}$ |
| $q_2$ | $\left\{ \frac{0.5}{q_4} \right\}$ | $\emptyset$ |
| $q_3$ | $\emptyset$ | $\left\{ \frac{0.7}{q_2} + \frac{0.1}{q_4}, \frac{0.1}{q_3} + \frac{0.7}{q_4} \right\}$ |
| $q_4$ | $\emptyset$ | $\emptyset$ |

Table 1: A fuzzy transition function.

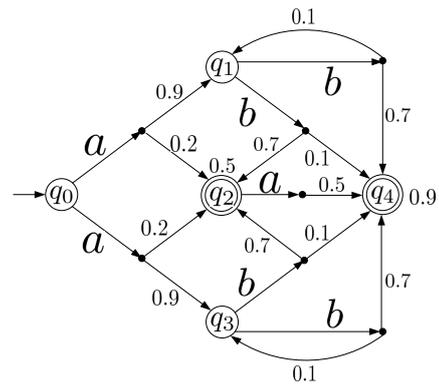

Figure 1: The nondeterministic fuzzy automaton.

We are now in the position to define the fuzzy language accepted by a nondeterministic fuzzy automaton. To this end, we need to extend the fuzzy transition function to strings.

Before stating the definition, let us fix a notational convention: if $\Phi \in \delta(q, a)$, then we identify $\delta(q, a)$ with $\delta(q, a) - \{\Phi\}$, since $\Phi$ has no contribution to the fuzzy language.

**Definition 4.** *Let $M = (Q, \Sigma, \delta, q_0, F)$ be a nondeterministic fuzzy automaton.*

(1) *The extended fuzzy transition function from $Q \times \Sigma^*$ to $\mathcal{P}(\mathcal{F}(Q))$, denoted by the same notation $\delta$, is defined inductively as follows:*

$$
\begin{aligned}
\delta(q, \epsilon) &= \left\{ \frac{1}{q} \right\} \\
\delta(q, sa) &= \left\{ \mu_s(p) \cdot \mu_p \,\middle|\, \mu_s \in \delta(q, s), p \in Q, \mu_p \in \delta(p, a) \right\}
\end{aligned}
$$

*for all $q \in Q$, $s \in \Sigma^*$, and $a \in \Sigma$.*

(2) *The* language *$\mathcal{L}(M)$ accepted by $M$ is a fuzzy subset of $\Sigma^*$ with the membership function defined by*

$$
\mathcal{L}(M)(s) = \vee \left\{ \mathrm{height}(\mu \cap F) \,\middle|\, \mu \in \delta(q_0, s) \right\}
$$

*for all $s \in \Sigma^*$. The membership $\mathcal{L}(M)(s)$ is the degree to which $s$ is accepted by $M$.*

Following the notion of fuzzy languages for deterministic fuzzy automata, max-min operator has been adopted in the above definition. There is no difficulty to check that this definition is consistent with that of fuzzy languages for deterministic fuzzy automata in the sense that viewing a deterministic fuzzy automaton $M$ as a deterministic fuzzy automaton and using the above definition to compute $\mathcal{L}(M)$ give the same result as directly using the definition of fuzzy languages for deterministic fuzzy automata.

**Example 2.** *Let us revisit Example 1. By definition, we have that*

$$
\begin{aligned}
\delta(q_0, \epsilon) &= \left\{ \frac{1}{q_0} \right\}, \\
\delta(q_0, a) &= \left\{ \mu_\epsilon(p) \cdot \mu_p \,\middle|\, \mu_\epsilon \in \delta(q_0, \epsilon), p \in Q, \mu_p \in \delta(p, a) \right\} \\
&= \left\{ \frac{0.9}{q_1} + \frac{0.2}{q_2}, \frac{0.2}{q_2} + \frac{0.9}{q_3} \right\}, \\
\delta(q_0, aa) &= \left\{ \mu_a(p) \cdot \mu_p \,\middle|\, \mu_a \in \delta(q_0, a), p \in Q, \mu_p \in \delta(p, a) \right\} \\
&= \left\{ \frac{0.2}{q_4} \right\}, \\
\delta(q_0, ab) &= \left\{ \mu_a(p) \cdot \mu_p \,\middle|\, \mu_a \in \delta(q_0, a), p \in Q, \mu_p \in \delta(p, b) \right\} \\
&= \left\{ \frac{0.1}{q_1} + \frac{0.7}{q_4}, \frac{0.7}{q_2} + \frac{0.1}{q_4}, \frac{0.1}{q_3} + \frac{0.7}{q_4} \right\}.
\end{aligned}
$$

*Therefore,*

$$
\begin{aligned}
\mathcal{L}(M)(\epsilon) &= \vee \{ \mathrm{height}(\mu \cap F) \,|\, \mu \in \delta(q_0, \epsilon) \} = 0, \\
\mathcal{L}(M)(a) &= \vee \{ \mathrm{height}(\mu \cap F) \,|\, \mu \in \delta(q_0, a) \} = 0.2, \\
\mathcal{L}(M)(aa) &= \vee \{ \mathrm{height}(\mu \cap F) \,|\, \mu \in \delta(q_0, aa) \} = 0.2, \\
\mathcal{L}(M)(ab) &= \vee \{ \mathrm{height}(\mu \cap F) \,|\, \mu \in \delta(q_0, ab) \} = 0.7.
\end{aligned}
$$

We end this section with a lemma, which will be needed later on.

**Lemma 1.** *Let $M = (Q, \Sigma, \delta, q_0, F)$ be a nondeterministic fuzzy automaton. Then*

$$
\mathcal{L}(M)(s) = \vee \{ \mathrm{height}(\mu \cap F) \,|\, \mu \in \delta(q_0, s) \} = \mathrm{height} \left[ \left( \cup_{\mu \in \delta(q_0, s)} \mu \right) \cap F \right]
$$

*for all $s \in \Sigma^*$.*

*Proof.* For any given $s \in \Sigma^*$, assume that $\delta(q_0, s) = \{\mu_i \mid i \in I\}$. Then we only need to verify that

$$\vee \{\text{height}(\mu_i \cap F) \mid i \in I\} = \text{height}\left[(\cup_{i \in I} \mu_i) \cap F\right]. \tag{1}$$

By definition, we have that

$$
\begin{aligned}
\vee \{\text{height}(\mu_i \cap F) \mid i \in I\} &= \vee_{i \in I} \text{height}(\mu_i \cap F) \\
&= \vee_{i \in I} \vee_{q \in Q} \left[\mu_i(q) \wedge F(q)\right] \\
&= \vee_{q \in Q} \vee_{i \in I} \left[\mu_i(q) \wedge F(q)\right] \\
&= \vee_{q \in Q} \{[\vee_{i \in I} \mu_i(q)] \wedge F(q)\} \\
&= \vee_{q \in Q} \{(\cup_{i \in I} \mu_i)(q) \wedge F(q)\} \\
&= \text{height}\left[(\cup_{i \in I} \mu_i) \cap F\right].
\end{aligned}
$$

This proves Eq. (1), as desired. $\qquad\square$

## 4. Nondeterministic fuzzy automata with $\epsilon$-moves

In this section, we extend the concepts of nondeterministic fuzzy automata and fuzzy languages accepted by them to nondeterministic fuzzy automata with $\epsilon$-moves. The notion of nondeterministic fuzzy automata with $\epsilon$-moves is also a generalization of nondeterministic finite automata with $\epsilon$-moves. Recall that $\epsilon$ represents the empty string. The $\epsilon$-move is sometimes called a *silent move* or *silent transition*. As for finite automata, $\epsilon$-moves can be regarded as internal transitions of a fuzzy automaton.

A nondeterministic fuzzy automaton with $\epsilon$-moves replaces the fuzzy transition function in a nondeterministic fuzzy automaton with one that allows the empty string $\epsilon$ as a possible input. More concretely, we have the following definition.

**Definition 5.** *A nondeterministic fuzzy automaton $M_\epsilon$ with $\epsilon$-moves is a five-tuple $(Q, \Sigma, \delta, q_0, F)$, where the components $Q, \Sigma, q_0, F$ are the same as those in Definition 3 and the fuzzy transition function $\delta$ is a function from $Q \times (\Sigma \cup \{\epsilon\})$ to $\mathcal{P}(\mathcal{F}(Q))$.*

We now turn our attention to the fuzzy language accepted by a nondeterministic fuzzy automaton with $\epsilon$-moves. Like Definition 4, we need to extend the fuzzy transition function to strings. To this end, it is convenient to introduce the following notations.

Let $M_\epsilon = (Q, \Sigma, \delta, q_0, F)$ be a nondeterministic fuzzy automaton with $\epsilon$-moves. Following Definition 4, we define a function $\delta_\epsilon : Q \times \{\epsilon\}^* \longrightarrow \mathcal{P}(\mathcal{F}(Q))$ as follows:

$$
\begin{aligned}
\delta_\epsilon(q, \epsilon) &= \left\{\frac{1}{q}\right\} \cup \delta(q, \epsilon) \\
\delta_\epsilon(q, \epsilon^n \epsilon) &= \left\{\mu_{\epsilon^n}(p) \cdot \mu_p \mid \mu_{\epsilon^n} \in \delta_\epsilon(q, \epsilon^n), p \in Q, \mu_p \in \delta_\epsilon(p, \epsilon)\right\}
\end{aligned}
$$

for all $q \in Q$ and $n \in \mathbb{N}$.

Further, for any $q \in Q$, we set

$$\Delta_\epsilon(q) = \cup_{n \in \mathbb{N}} \delta_\epsilon(q, \epsilon^n)$$

and call it the $\epsilon$-*closure of* $q$. For any $\mu \in \mathcal{F}(Q)$, the $\epsilon$-closure of $\mu$ is defined as

$$\Delta_\epsilon(\mu) = \{\mu\} \cup \{\mu(q) \cdot \eta \mid q \in Q, \eta \in \Delta_\epsilon(q)\}.$$

**Definition 6.** *Let $M_\epsilon = (Q, \Sigma, \delta, q_0, F)$ be a nondeterministic fuzzy automaton with $\epsilon$-moves.*

(1) *The* extended fuzzy transition function *from $Q \times \Sigma^*$ to $\mathcal{P}(\mathcal{F}(Q))$, denoted by $\widehat{\delta}$, is defined inductively as follows:*

$$
\begin{aligned}
\widehat{\delta}(q, \epsilon) &= \Delta_\epsilon(q) \\
\widehat{\delta}(q, sa) &= \cup \left\{\Delta_\epsilon\left(\mu_s(p) \cdot \mu_p\right) \mid \mu_s \in \widehat{\delta}(q, s), p \in Q, \mu_p \in \delta(p, a)\right\}
\end{aligned}
$$

*for all $q \in Q$, $s \in \Sigma^*$, and $a \in \Sigma$.*

(2) *The* language $\mathcal{L}(M_\epsilon)$ *accepted by* $M_\epsilon$ *is a fuzzy subset of* $\Sigma^*$ *with the membership function defined by*

$$\mathcal{L}(M_\epsilon)(s) = \vee \left\{ \text{height}(\mu \cap F) \, | \, \mu \in \widehat{\delta}(q_0, s) \right\}$$

*for all* $s \in \Sigma^*$.

Note that in general, it does not hold that $\widehat{\delta}(q, a) = \delta(q, a)$, because the $\epsilon$-moves have been involved. In the same way, it is not necessary that $\widehat{\delta}(q, \epsilon) = \delta(q, \epsilon)$.

We illustrate the above notions by an example.

| $\delta$ | $a$ | $b$ | $\epsilon$ |
|---|---|---|---|
| $q_0$ | $\left\{ \frac{0.9}{q_1} + \frac{0.2}{q_2}, \frac{0.2}{q_2} + \frac{0.9}{q_3} \right\}$ | $\emptyset$ | $\left\{ \frac{0.7}{q_2} \right\}$ |
| $q_1$ | $\emptyset$ | $\left\{ \frac{0.1}{q_1} + \frac{0.7}{q_4}, \frac{0.7}{q_2} + \frac{0.1}{q_4} \right\}$ | $\left\{ \frac{0.8}{q_4} \right\}$ |
| $q_2$ | $\left\{ \frac{0.5}{q_4} \right\}$ | $\emptyset$ | $\emptyset$ |
| $q_3$ | $\emptyset$ | $\left\{ \frac{0.7}{q_2} + \frac{0.1}{q_4}, \frac{0.1}{q_3} + \frac{0.7}{q_4} \right\}$ | $\left\{ \frac{0.5}{q_4} \right\}$ |
| $q_4$ | $\emptyset$ | $\emptyset$ | $\emptyset$ |

Table 2: The fuzzy transition function in Example 3.

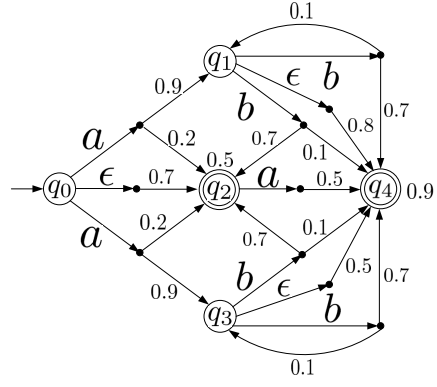

Figure 2: The nondeterministic fuzzy automaton with $\epsilon$-moves.

**Example 3.** *Let us augment the nondeterministic fuzzy automaton* $M$ *in Example 1 with* $\epsilon$-*moves. The corresponding fuzzy transition function is defined in Table 2 and the nondeterministic fuzzy automaton* $M_\epsilon$ *with* $\epsilon$-*moves is shown in Figure 2.*

*By definition, we see that*

$$\begin{aligned}
\delta_\epsilon(q_0, \epsilon) &= \left\{ \frac{1}{q_0}, \frac{0.7}{q_2} \right\}, \\
\delta_\epsilon(q_0, \epsilon\epsilon) &= \left\{ \mu_\epsilon(p) \cdot \mu_p \, | \, \mu_\epsilon \in \delta_\epsilon(q, \epsilon), p \in Q, \mu_p \in \delta_\epsilon(p, \epsilon) \right\} \\
&= \left\{ \frac{1}{q_0}, \frac{0.7}{q_2} \right\}.
\end{aligned}$$

*By the same token, we find that* $\delta_\epsilon(q_0, \epsilon^n) = \left\{ \frac{1}{q_0}, \frac{0.7}{q_2} \right\}$ *for any* $n \in \mathbb{N}$, *and thus,* $\Delta_\epsilon(q_0) = \left\{ \frac{1}{q_0}, \frac{0.7}{q_2} \right\}$. *Similarly, we get that* $\Delta_\epsilon(q_1) = \left\{ \frac{1}{q_1}, \frac{0.8}{q_4} \right\}$, $\Delta_\epsilon(q_2) = \left\{ \frac{1}{q_2} \right\}$, $\Delta_\epsilon(q_3) = \left\{ \frac{1}{q_3}, \frac{0.5}{q_4} \right\}$, *and* $\Delta_\epsilon(q_4) = \left\{ \frac{1}{q_4} \right\}$.

*Further, a routine computation shows that*

$$\begin{aligned}
\widehat{\delta}(q_0, \epsilon) &= \Delta_\epsilon(q_0) = \left\{ \frac{1}{q_0}, \frac{0.7}{q_2} \right\}, \\
\widehat{\delta}(q_0, a) &= \cup \left\{ \Delta_\epsilon \left( \mu_\epsilon(p) \cdot \mu_p \right) \, | \, \mu_\epsilon \in \widehat{\delta}(q_0, \epsilon), p \in Q, \mu_p \in \delta(p, a) \right\} \\
&= \Delta_\epsilon \left( \frac{0.9}{q_1} + \frac{0.2}{q_2} \right) \cup \Delta_\epsilon \left( \frac{0.2}{q_2} + \frac{0.9}{q_3} \right) \cup \Delta_\epsilon \left( \frac{0.5}{q_4} \right) \\
&= \left\{ \frac{0.9}{q_1} + \frac{0.2}{q_2}, \frac{0.2}{q_2} + \frac{0.9}{q_3}, \frac{0.9}{q_1}, \frac{0.2}{q_2}, \frac{0.9}{q_3}, \frac{0.5}{q_4}, \frac{0.8}{q_4} \right\}.
\end{aligned}$$

*Similarly, we can obtain that*

$$\widehat{\delta}(q_0, b) = \widehat{\delta}(q_1, a) = \widehat{\delta}(q_2, b) = \widehat{\delta}(q_3, a) = \widehat{\delta}(q_4, a) = \widehat{\delta}(q_4, b) = \emptyset,$$

$$\widehat{\delta}(q_1, b) = \left\{ \frac{0.1}{q_1} + \frac{0.7}{q_4}, \frac{0.7}{q_2} + \frac{0.1}{q_4}, \frac{0.1}{q_1}, \frac{0.7}{q_2}, \frac{0.1}{q_4}, \frac{0.7}{q_4} \right\},$$

$$\widehat{\delta}(q_2, a) = \left\{ \frac{0.5}{q_4} \right\},$$

$$\widehat{\delta}(q_3, b) = \left\{ \frac{0.7}{q_2} + \frac{0.1}{q_4}, \frac{0.1}{q_3} + \frac{0.7}{q_4}, \frac{0.7}{q_2}, \frac{0.1}{q_3}, \frac{0.1}{q_4}, \frac{0.7}{q_4} \right\}.$$

*Hence,*

$$
\begin{aligned}
\mathcal{L}(M_\epsilon)(\epsilon) &= \vee \left\{ \text{height}(\mu \cap F) \,|\, \mu \in \widehat{\delta}(q_0, \epsilon) \right\} = 0.5, \\
\mathcal{L}(M_\epsilon)(a) &= \vee \left\{ \text{height}(\mu \cap F) \,|\, \mu \in \widehat{\delta}(q_0, a) \right\} = 0.8, \\
\mathcal{L}(M_\epsilon)(b) &= \vee \left\{ \text{height}(\mu \cap F) \,|\, \mu \in \widehat{\delta}(q_0, b) \right\} = 0.
\end{aligned}
$$

## 5. Equivalences

It is well known that for any nondeterministic finite automaton with or without $\epsilon$-moves, there exists an equivalent deterministic finite automaton such that both of them accept the same language [15]. In this section, we would like to extend this equivalence to fuzzy automata.

Let us first establish the following theorem, which shows that nondeterministic fuzzy automata without $\epsilon$-moves and deterministic fuzzy automata accept the same class of fuzzy languages.

**Theorem 1.** *For any nondeterministic fuzzy automaton $M$, there exists a deterministic fuzzy automaton $M'$ such that $\mathcal{L}(M) = \mathcal{L}(M')$.*

*Proof.* Let $M = (Q, \Sigma, \delta, q_0, F)$. We construct $M' = (Q', \Sigma', \delta', q_0', F')$ as follows: All the components are the same as those in $M$ but the fuzzy transition function $\delta' : Q' \times \Sigma' \longrightarrow \mathcal{F}(Q')$, which is defined by

$$
\delta'(q, a) = \begin{cases}
\cup_{\mu \in \delta(q, a)} \mu, & \text{if } \delta(q, a) \neq \emptyset \\[2mm]
\Phi, & \text{otherwise}
\end{cases}
$$

for any $q \in Q = Q'$ and $a \in \Sigma = \Sigma'$. Clearly, this construction gives rise to a deterministic fuzzy automaton $M'$. It remains to check that $\mathcal{L}(M) = \mathcal{L}(M')$. By Lemma 1, for any $s \in \Sigma^*$ we have that

$$\mathcal{L}(M)(s) = \text{height}\left[ \left( \cup_{\mu \in \delta(q_0, s)} \mu \right) \cap F \right].$$

On the other hand, it follows from Definition 2 that for any $s \in \Sigma^*$,

$$\mathcal{L}(M')(s) = \text{height}\left( \delta'(q_0, s) \cap F' \right) = \text{height}\left( \delta'(q_0, s) \cap F \right).$$

Therefore, to prove $\mathcal{L}(M) = \mathcal{L}(M')$, it is sufficient to show that

$$\delta'(q, s) = \cup_{\mu \in \delta(q, s)} \mu \tag{2}$$

for any $q \in Q = Q'$ and $s \in \Sigma^* = \Sigma'^*$. This will be proved by induction on the length of the string $s$.

The basis step is for strings of length 0, namely, $s = \epsilon$. In this case, it follows from definition that $\delta(q, \epsilon) = \{1/q\}$ and thus

$$\delta'(q, \epsilon) = \frac{1}{q} = \cup_{\mu \in \delta(q, \epsilon)} \mu.$$

Therefore, the basis step is true.

The induction hypothesis is that Eq. (2) holds for all strings $s$ with length $n$. We now show that for any $a \in \Sigma = \Sigma'$,

$$\delta'(q, sa) = \cup_{\mu \in \delta(q, sa)} \mu. \tag{3}$$

In fact, we get by Definition 2 and the induction hypothesis that

$$
\begin{aligned}
\delta'(q, sa) &= \cup_{p \in Q} \left[ \delta'(q, s)(p) \cdot \delta'(p, a) \right] \\
&= \cup_{p \in Q} \left[ \left( \cup_{\mu_s \in \delta(q, s)} \mu_s \right)(p) \cdot \delta'(p, a) \right] \\
&= \cup_{p \in Q} \left[ \left( \vee_{\mu_s \in \delta(q, s)} \mu_s(p) \right) \cdot \delta'(p, a) \right] \\
&= \cup_{p \in Q} \left[ \cup_{\mu_s \in \delta(q, s)} \mu_s(p) \cdot \delta'(p, a) \right] \\
&= \cup_{p \in Q} \cup_{\mu_s \in \delta(q, s)} \left[ \mu_s(p) \cdot \left( \cup_{\mu_p \in \delta(p, a)} \mu_p \right) \right] \\
&= \cup_{p \in Q} \cup_{\mu_s \in \delta(q, s)} \left[ \cup_{\mu_p \in \delta(p, a)} \left( \mu_s(p) \cdot \mu_p \right) \right] \\
&= \cup_{\mu_s \in \delta(q, s)} \cup_{p \in Q} \cup_{\mu_p \in \delta(p, a)} \left( \mu_s(p) \cdot \mu_p \right) \\
&= \cup \left\{ \mu_s(p) \cdot \mu_p \mid \mu_s \in \delta(q, s), p \in Q, \mu_p \in \delta(p, a) \right\} \\
&= \cup \delta(q, sa) \\
&= \cup_{\mu \in \delta(q, sa)} \mu.
\end{aligned}
$$

Hence, Eq. (3) holds. This completes the proof of the theorem. $\square$

As an example, let us convert the nondeterministic fuzzy automaton in Example 1 into a deterministic one by the construction introduced in the above proof.

**Example 4.** *Consider the nondeterministic fuzzy automaton $M$ in Example 1. Following the construction in the proof of Theorem 1, we set $M' = (Q, \Sigma, \delta', q_0, F)$, where $\delta'$ is defined in Table 3 and the transition diagram of $M'$ is depicted in Figure 3.*

| $\delta'$ | $a$ | $b$ |
|---|---|---|
| $q_0$ | $\frac{0.9}{q_1} + \frac{0.2}{q_2} + \frac{0.9}{q_3}$ | $\Phi$ |
| $q_1$ | $\Phi$ | $\frac{0.1}{q_1} + \frac{0.7}{q_2} + \frac{0.7}{q_4}$ |
| $q_2$ | $\frac{0.5}{q_4}$ | $\Phi$ |
| $q_3$ | $\Phi$ | $\frac{0.7}{q_2} + \frac{0.1}{q_3} + \frac{0.7}{q_4}$ |
| $q_4$ | $\Phi$ | $\Phi$ |

Table 3: The fuzzy transition function in Example 4.

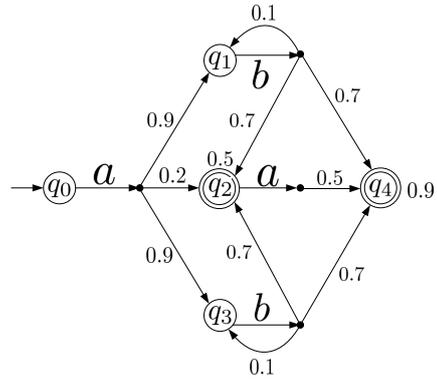

Figure 3: The deterministic fuzzy automaton.

To establish a theorem like Theorem 1 for nondeterministic fuzzy automata with $\epsilon$-moves, we need two lemmas.

**Lemma 2.** *Let $M_\epsilon = (Q, \Sigma, \delta, q_0, F)$ be a nondeterministic fuzzy automaton with $\epsilon$-moves. Then*

$$\mathcal{L}(M_\epsilon)(s) = \vee \left\{ \text{height}(\mu \cap F) \mid \mu \in \widehat{\delta}(q_0, s) \right\} = \text{height} \left[ \left( \cup_{\mu \in \widehat{\delta}(q_0, s)} \mu \right) \cap F \right]$$

*for all $s \in \Sigma^*$.*

*Proof.* It is similar to that of Lemma 1, and thus we omit it here. □

The following lemma, which gives another expression of the extended fuzzy transition function, is useful to prove the next theorem.

**Lemma 3.** *Let $M_\epsilon = (Q, \Sigma, \delta, q_0, F)$ be a nondeterministic fuzzy automaton with $\epsilon$-moves. Then*

$$\widehat{\delta}(q, sa) = \cup \left\{ \Delta_\epsilon \left( \mu_s(p) \cdot \mu_p \right) \big| \mu_s \in \widehat{\delta}(q, s), p \in Q, \mu_p \in \delta(p, a) \right\} = \left\{ \mu_s(p) \cdot \mu_p \big| \mu_s \in \widehat{\delta}(q, s), p \in Q, \mu_p \in \widehat{\delta}(p, a) \right\} \quad (4)$$

*for any $q \in Q$, $s \in \Sigma^*$, and $a \in \Sigma$.*

*Proof.* To prove Eq. (4), it suffices to verify that

$$\cup_{\mu_s \in \widehat{\delta}(q,s)} \cup_{p \in Q} \cup_{\mu_p \in \delta(p,a)} \Delta_\epsilon \left( \mu_s(p) \cdot \mu_p \right) = \cup_{\mu_s \in \widehat{\delta}(q,s)} \cup_{p \in Q} \cup_{\mu_p \in \widehat{\delta}(p,a)} \left\{ \mu_s(p) \cdot \mu_p \right\}. \quad (5)$$

Let $\alpha \in \cup_{\mu_s \in \widehat{\delta}(q,s)} \cup_{p \in Q} \cup_{\mu_p \in \delta(p,a)} \Delta_\epsilon \left( \mu_s(p) \cdot \mu_p \right)$. Then there exist $\mu_1 \in \widehat{\delta}(q, s)$, $p' \in Q$, and $\mu_2 \in \delta(p', a)$ such that

$$\alpha \in \Delta_\epsilon \left( \mu_1(p') \cdot \mu_2 \right) = \left\{ \mu_1(p') \right\} \cup \left\{ \left[ \mu_1(p') \wedge \mu_2(q') \right] \cdot \eta \, | \, q' \in Q, \eta \in \Delta_\epsilon(q') \right\}.$$

Therefore, we see that either $\alpha = \mu_1(p') \cdot \mu_2$, or $\alpha = \left[ \mu_1(p') \wedge \mu_2(q') \right] \cdot \eta$ for some $q' \in Q$ and $\eta \in \delta_\epsilon(q', \epsilon^k) \subseteq \Delta_\epsilon(q')$. For the first case, it follows immediately that $\alpha$ belongs to the right of Eq. (5). For the second case, take $\mu_s = \mu_1$, $p = p'$, and $\mu_p = \mu_2(q') \cdot \eta$. Noting that $\mu_p = \mu_2(q') \cdot \eta \in \widehat{\delta}(p, a)$, we also find that $\alpha$ belongs to the right of Eq. (5). Consequently, $\cup_{\mu_s \in \widehat{\delta}(q,s)} \cup_{p \in Q} \cup_{\mu_p \in \delta(p,a)} \Delta_\epsilon \left( \mu_s(p) \cdot \mu_p \right) \subseteq \cup_{\mu_s \in \widehat{\delta}(q,s)} \cup_{p \in Q} \cup_{\mu_p \in \widehat{\delta}(p,a)} \left\{ \mu_s(p) \cdot \mu_p \right\}$.

Conversely, suppose that $\alpha$ belongs to the right of Eq. (5). Then there are $\mu_1 \in \widehat{\delta}(q, s)$, $p' \in Q$, and $\mu_2 \in \widehat{\delta}(p', a)$ such that $\alpha = \mu_1(p') \cdot \mu_2$. Because $\widehat{\delta}(p', a) = \cup_{\mu_\epsilon \in \widehat{\delta}(p', \epsilon)} \cup_{r \in Q} \cup_{\mu_r \in \delta(r,a)} \Delta_\epsilon \left( \mu_\epsilon(r) \cdot \mu_r \right)$, there exist $\mu_\epsilon \in \widehat{\delta}(p', \epsilon)$, $r \in Q$, and $\mu_r \in \delta(r, a)$ such that

$$\mu_2 \in \Delta_\epsilon \left( \mu_\epsilon(r) \cdot \mu_r \right) = \left\{ \mu_\epsilon(r) \cdot \mu_r \right\} \cup \left\{ \left[ \mu_\epsilon(r) \wedge \mu_r(q') \right] \cdot \eta \, | \, q' \in Q, \eta \in \Delta_\epsilon(q') \right\}.$$

We thus get that either $\mu_2 = \mu_\epsilon(r) \cdot \mu_r$, or $\mu_2 = \left[ \mu_\epsilon(r) \wedge \mu_r(q') \right] \cdot \eta$ for some $q' \in Q$ and $\eta \in \delta_\epsilon(q', \epsilon^k) \subseteq \Delta_\epsilon(q')$. For both of the cases, take $\mu_s = \mu_1(p') \cdot \mu_\epsilon$, $p = r$, and $\mu_p = \mu_r$. Since $\mu_s = \mu_1(p') \cdot \mu_\epsilon \in \widehat{\delta}(q, s)$, we see that $\alpha$ belongs to the left of Eq. (5) in either case. Hence, Eq. (5) holds, finishing the proof of the lemma. □

With the above lemmas, we can now show that both nondeterministic fuzzy automata with $\epsilon$-moves and nondeterministic fuzzy automata without $\epsilon$-moves accept the same class of fuzzy languages.

**Theorem 2.** *For any nondeterministic fuzzy automaton $M_\epsilon$ with $\epsilon$-moves, there exists a nondeterministic fuzzy automaton $M'$ such that $\mathcal{L}(M_\epsilon) = \mathcal{L}(M')$.*

*Proof.* Let $M_\epsilon = (Q, \Sigma, \delta, q_0, F)$. We construct a nondeterministic fuzzy automaton $M' = (Q', \Sigma', \delta', q_0', F')$ as follows: $Q' = Q$, $\Sigma' = \Sigma$, $q_0' = q_0$, and $\delta' : Q' \times \Sigma' \longrightarrow \mathcal{P}(\mathcal{F}(Q'))$ and $F'$ are defined by

$$\begin{aligned} \delta'(q, a) &= \widehat{\delta}(q, a), \\ F'(q) &= \text{height} \left[ \left( \cup_{\mu \in \Delta_\epsilon(q)} \mu \right) \cap F \right] \end{aligned}$$

for any $q \in Q' = Q$ and $a \in \Sigma' = \Sigma$, where $\widehat{\delta}$ is the extended fuzzy transition function introduced in Definition 6. Clearly, $M'$ is a nondeterministic fuzzy automaton without $\epsilon$-moves. It remains to check that $\mathcal{L}(M_\epsilon) = \mathcal{L}(M')$, i.e., $\mathcal{L}(M_\epsilon)(s) = \mathcal{L}(M')(s)$ for every $s \in \Sigma^*$. We prove it by induction on the length of $s$.

For $s = \epsilon$, we get by Lemma 2 that

$$\mathcal{L}(M_\epsilon)(\epsilon) = \text{height} \left[ \left( \cup_{\mu \in \widehat{\delta}(q_0, \epsilon)} \mu \right) \cap F \right] = \text{height} \left[ \left( \cup_{\mu \in \Delta_\epsilon(q_0)} \mu \right) \cap F \right].$$

On the other hand, it follows from Lemma 1 and the construction of $M'$ that

$$\mathcal{L}(M')(\epsilon) = \text{height} \left[ \left( \cup_{\mu \in \delta'(q_0, \epsilon)} \mu \right) \cap F' \right] = F'(q_0) = \text{height} \left[ \left( \cup_{\mu \in \Delta_\epsilon(q_0)} \mu \right) \cap F \right].$$

Whence, we obtain that $\mathcal{L}(M_\epsilon)(\epsilon) = \mathcal{L}(M')(\epsilon)$, which establishes the basis step.

For the induction step, suppose that $\mathcal{L}(M_\epsilon)(s) = \mathcal{L}(M')(s)$ for every $s \in \Sigma^*$ with length less than or equal to $n$. Then it follows from Lemmas 1 and 2 that

$$\vee_{z \in Q} \vee_{\mu_z \in \widehat{\delta}(q_0, s)} [\mu_s(z) \wedge F(z)] = \vee_{z \in Q} \vee_{\mu_s \in \delta'(q_0, s)} \vee_{\eta \in \Delta_\epsilon(z)} \vee_{y \in Q} [\mu_s(z) \wedge \eta(y) \wedge F(y)]. \tag{6}$$

To consider $sa$, we first prove the following claim: For any $q \in Q$ and $w \in \Sigma^*$ with $w \neq \epsilon$,

$$\delta'(q, w) = \widehat{\delta}(q, w). \tag{7}$$

Again, we use induction on the length of $w$. In the basis step, namely, $w \in \Sigma$, the claim follows directly from the definition of $\delta'$. The induction hypothesis is that $\delta'(q, w) = \widehat{\delta}(q, w)$ for all strings $w$ with length less than or equal to $n$. We now prove the same for strings of the form $wa$. By Definition 4, we have that

$$\delta'(q, wa) = \left\{ \mu_w(p) \cdot \mu_p \,\middle|\, \mu_w \in \delta'(q, w), p \in Q, \mu_p \in \delta'(p, a) \right\},$$

and by Lemma 3 we see that

$$\widehat{\delta}(q, wa) = \left\{ \mu_w(p) \cdot \mu_p \,\middle|\, \mu_w \in \widehat{\delta}(q, w), p \in Q, \mu_p \in \widehat{\delta}(p, a) \right\}.$$

Therefore, we obtain by the basis step and the induction hypothesis that $\delta'(q, wa) = \widehat{\delta}(q, wa)$. This proves the claim.

We are ready to verify that $\mathcal{L}(M_\epsilon)(sa) = \mathcal{L}(M')(sa)$. By using Lemmas 1 and 2 and Eq. (6), we have that

$$
\begin{aligned}
\mathcal{L}(M')(sa) &= \vee_{z \in Q} \vee_{\mu \in \delta'(q_0, sa)} [\mu(z) \wedge F'(z)] \\
&= \vee_{z \in Q} \vee_{\mu \in \delta'(q_0, sa)} \left\{ \mu(z) \wedge \left[ \vee_{y \in Q} \vee_{\eta \in \Delta_\epsilon(z)} [\eta(y) \wedge F(y)] \right] \right\} \\
&= \vee_{z \in Q} \vee_{\mu \in \delta'(q_0, sa)} \vee_{y \in Q} \vee_{\eta \in \Delta_\epsilon(z)} [\mu(z) \wedge \eta(y) \wedge F(y)] \\
&= \vee_{z \in Q} \vee_{\mu_s \in \delta'(q_0, s)} \vee_{p \in Q} \vee_{\mu_p \in \delta'(p, a)} \vee_{y \in Q} \vee_{\eta \in \Delta_\epsilon(z)} [\mu_s(p) \wedge \mu_p(z) \wedge \eta(y) \wedge F(y)] \\
&= \vee_{z \in Q} \vee_{\mu_s \in \delta'(q_0, s)} \vee_{p \in Q} \vee_{\mu_p \in \delta'(p, a)} [\mu_s(p) \wedge \mu_p(z) \wedge F(z)] \\
&= \vee_{z \in Q} \vee_{\mu_s \in \widehat{\delta}(q_0, s)} \vee_{p \in Q} \vee_{\mu_p \in \widehat{\delta}(p, a)} [\mu_s(p) \wedge \mu_p(z) \wedge F(z)] \\
&= \vee_{z \in Q} \vee_{\mu \in \widehat{\delta}(q_0, sa)} [\mu(z) \wedge F(z)] \\
&= \mathcal{L}(M_\epsilon)(sa),
\end{aligned}
$$

namely, $\mathcal{L}(M_\epsilon)(sa) = \mathcal{L}(M')(sa)$. This completes the proof of the theorem. $\qquad\square$

To illustrate the construction in the above theorem, we examine an example.

**Example 5.** *Consider the nondeterministic fuzzy automaton $M_\epsilon$ with $\epsilon$-moves in Example 3. Using the construction in the proof of Theorem 2, the corresponding nondeterministic fuzzy automaton is $M' = (Q, \Sigma, \delta', q_0, F')$, where $\delta'$ is specified by Table 4 and $F'$ is defined as follows:*

$$
\begin{aligned}
F'(q_0) &= \mathrm{height}\left[ \left( \cup_{\mu \in \Delta_\epsilon(q_0)} \mu \right) \cap F \right] = 0.5, \\
F'(q_1) &= \mathrm{height}\left[ \left( \cup_{\mu \in \Delta_\epsilon(q_1)} \mu \right) \cap F \right] = 0.8, \\
F'(q_2) &= \mathrm{height}\left[ \left( \cup_{\mu \in \Delta_\epsilon(q_2)} \mu \right) \cap F \right] = 0.5, \\
F'(q_3) &= \mathrm{height}\left[ \left( \cup_{\mu \in \Delta_\epsilon(q_3)} \mu \right) \cap F \right] = 0.5, \\
F'(q_4) &= \mathrm{height}\left[ \left( \cup_{\mu \in \Delta_\epsilon(q_4)} \mu \right) \cap F \right] = 0.9.
\end{aligned}
$$

*Since the transition diagram of $M'$ is complicated, we omit it here.*

Summarizing Theorems 1 and 2, we get immediately the following corollary.

**Corollary 1.** *Let $\mathcal{L}$ be a fuzzy language. Then the following statements are equivalent:*

(1) *There exists a deterministic fuzzy automaton $M$ such that $\mathcal{L}(M) = \mathcal{L}$.*

(2) *There exists a nondeterministic fuzzy automaton $M'$ such that $\mathcal{L}(M') = \mathcal{L}$.*

(3) *There exists a nondeterministic fuzzy automaton $M_\epsilon$ with $\epsilon$-moves such that $\mathcal{L}(M_\epsilon) = \mathcal{L}$.*

| $\delta'$ | $a$ | $b$ |
|---|---|---|
| $q_0$ | $\left\{\frac{0.9}{q_1} + \frac{0.2}{q_2}, \frac{0.2}{q_2} + \frac{0.9}{q_3}, \frac{0.9}{q_1}, \frac{0.2}{q_2}, \frac{0.9}{q_3}, \frac{0.5}{q_4}, \frac{0.8}{q_4}\right\}$ | $\emptyset$ |
| $q_1$ | $\emptyset$ | $\left\{\frac{0.1}{q_1} + \frac{0.7}{q_4}, \frac{0.7}{q_4} + \frac{0.1}{q_4}, \frac{0.1}{q_1}, \frac{0.7}{q_2}, \frac{0.1}{q_4}, \frac{0.7}{q_4}\right\}$ |
| $q_2$ | $\left\{\frac{0.5}{q_4}\right\}$ | $\emptyset$ |
| $q_3$ | $\emptyset$ | $\left\{\frac{0.7}{q_2} + \frac{0.1}{q_4}, \frac{0.1}{q_3} + \frac{0.7}{q_4}, \frac{0.7}{q_2}, \frac{0.1}{q_3}, \frac{0.1}{q_4}, \frac{0.7}{q_4}\right\}$ |
| $q_4$ | $\emptyset$ | $\emptyset$ |

Table 4: The fuzzy transition function of $M'$ in Example 5.

## 6. Conclusion

In the paper, we have investigated fuzzy automata with nondeterminism by introducing nondeterministic fuzzy automata and nondeterministic fuzzy automata with $\epsilon$-moves. Like the relationship among deterministic finite automata, nondeterministic finite automata, and nondeterministic finite automata with $\epsilon$-moves, we have shown that deterministic fuzzy automata, nondeterministic fuzzy automata, and nondeterministic fuzzy automata with $\epsilon$-moves are equivalent in the sense that they accept the same class of fuzzy languages. As an application of nondeterministic fuzzy automata, we are currently using them to model and specify nondeterministic fuzzy discrete-event systems.

## Acknowledgements

This work was supported by the National Natural Science Foundation of China under Grants 60873061, 60973004, and 70890080 and by the National Basic Research Program of China (973 Program) under Grants 2007CB311003, 2009CB320701, and 2010CB328103.